\documentclass{article}

\PassOptionsToPackage{numbers, compress}{natbib}
\usepackage[final]{neurips_2024}

\usepackage[utf8]{inputenc} 
\usepackage[T1]{fontenc}    
\usepackage{hyperref}       
\usepackage{url}            
\usepackage{booktabs}       
\usepackage{amsfonts}       
\usepackage{nicefrac}       
\usepackage{microtype}      
\usepackage{xcolor}         
\usepackage{amsmath}
\usepackage{graphicx}
\usepackage{multirow}
\usepackage{float}

\newcommand*\samethanks[1][\value{footnote}]{\footnotemark[#1]}

\title{Towards Effective GenAI Multi-Agent Collaboration: \\Design and Evaluation for Enterprise Applications}

\author{Raphael Shu\thanks{Authors contributed equally},\enspace Nilaksh Das\samethanks,\enspace Michelle Yuan\samethanks,\enspace Monica Sunkara,\enspace Yi Zhang\\
  AWS Bedrock \\
}

\begin{document}

\maketitle

\begin{abstract}
AI agents powered by large language models (LLMs) have shown strong capabilities in problem solving. Through combining many intelligent agents, multi-agent collaboration has emerged as a promising approach to tackle complex, multi-faceted problems that exceed the capabilities of single AI agents. However, designing the collaboration protocols and evaluating the effectiveness of these systems remains a significant challenge, especially for enterprise applications. This report addresses these challenges by presenting a comprehensive evaluation of coordination and routing capabilities in a novel multi-agent collaboration framework. We evaluate two key operational modes: (1) a coordination mode enabling complex task completion through parallel communication and payload referencing, and (2) a routing mode for efficient message forwarding between agents. We benchmark on a set of handcrafted scenarios from three enterprise domains, which are publicly released with the report. For coordination capabilities, we demonstrate the effectiveness of inter-agent communication and payload referencing mechanisms, achieving end-to-end goal success rates of 90\%.
 Our analysis yields several key findings: multi-agent collaboration enhances goal success rates by up to 70\% compared to single-agent approaches in our benchmarks; payload referencing improves performance on code-intensive tasks by 23\%; latency can be substantially reduced with a routing mechanism that selectively bypasses agent orchestration. These findings offer valuable guidance for enterprise deployments of multi-agent systems and advance the development of scalable, efficient multi-agent collaboration frameworks.

\end{abstract}
\section{Introduction}
\label{sec:intro}

The rapid advancement of AI agents driven by large language models (LLMs)~\cite{gpt3} has opened new frontiers towards solving complex problems. Based on the strong reasoning and tool-use capabilities, an agent can plan and execute multiple steps for actions until the goal of problem solving is reached~\cite{yao2023react}. However, as the complexity of real-world challenges continues to grow, there is an increasing need for scaling up the agent-based systems by coordinating with multiple agents with diverse specializations~\cite{chatdev}.

Towards tackling multi-faceted real-world problems, \emph{multi-agent system (MAS)} research emerged in the mid-1980s to early-1990s as a critical sub-field of artificial intelligence focused on developing computational systems composed of multiple interacting intelligent agents~\cite{singh1994multiagent}. Researchers sought to create frameworks where autonomous entities could communicate, coordinate, and solve problems collectively~\cite{sycara1998multiagent}. With the rise of LLM-based AI agents in recent years, the key challenges in MAS research
regained focal attention in the new Generative AI (GenAI) era~\cite{han2024llm}. While earlier MAS work drew inspirations heavily from fields like distributed computing and game theory, new LLM-based GenAI agent research looks further into inter-disciplinary influence from psychology and social science as the AI agents start to demonstrate human-like intelligence as well as social behavior~\cite{Park2023GenerativeAgents}. 

One particularly fruitful research avenue in GenAI MAS research is the exploration of \emph{multi-agent collaboration (MAC)}~\cite{hong2024metagpt}. Operating under the ``collaborative assumption''~\cite{li-etal-2023-theory} -- a premise that agents are fundamentally motivated to achieve shared or compatible goals and prioritize collective problem solving over individual self-interest -- multi-agent collaboration aims to address the key challenges such as communication protocols, goal alignment, group decision-making, scalability, and trust and reliability~\cite{chen2023agentverse,talebirad2023multi,zhang2024towards}. 
In this paper, we present a particular design of MAC framework targeting the development of MAC applications for real-world enterprise use cases. Three research questions arise for having multiple LLM-based agents to work together: 1) how to define the collaboration mechanism, 2) how to facilitate efficient knowledge exchange between agents, and 3) how to evaluate the effectiveness and efficiency of collaboration. In this technical report, we propose strategies to address these research questions. Based on collected evaluation datasets, we report the evaluation results and detailed analysis on the effectiveness and efficiency of multi-agent collaboration.

\paragraph{Collaboration Mechanism} Collaboration mechanisms define how agents interact with each other to achieve a common goal~\cite{zhang-etal-2024-exploring}. Two key aspects are important: team hierarchy and decision-making mechanisms. Both hierarchy and decision-making can be either centralized or decentralized~\cite{guo2024large}. In a centralized hierarchy, a central authority assigns or delegates tasks to agents, whereas in a decentralized hierarchy, agents take their own initiatives. On the other hand, centralized decision-making involves a single agent making the final decision, while decentralized decision-making often relies on consensus or voting among multiple agents.

The literature presents various approaches to these mechanisms. For example, ChatDev~\cite{chatdev} employs a centralized team hierarchy and decision-making mechanism, while MAD~\cite{mad_2024} and Generative Agents~\cite{Park2023GenerativeAgents} adopt decentralized hierarchies and decision-making processes.

\begin{figure}[t]
    \centering
    \includegraphics[width=0.7\textwidth]{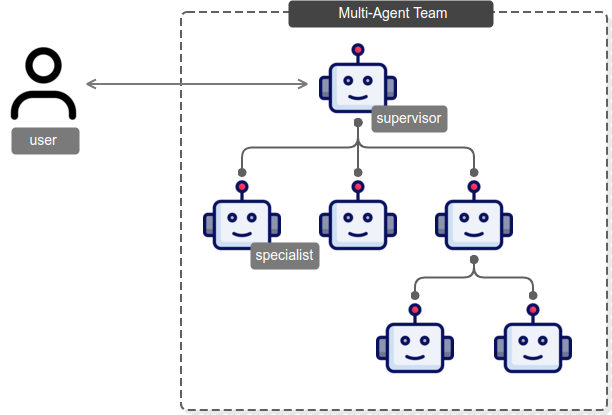}
    \caption{Illustration of the hierarchical agents approach for multi-agent collaboration. In a centralized hierarchy, a supervisor agent oversees and assigns tasks to specialist agents. The figure demonstrates a multi-layer hierarchy, where an agent can function as both a specialist agent and a supervisor agent.}
    \label{fig:overview}
\end{figure}

In MAC, we start by exploring centralized approaches which we refer to as hierarchical agents, and expand to decentralized mechanisms as future work. In hierarchical agents, each team has a tree-like hierarchy, with the root agent responsible for the the team goal and the "leaf" agents responsible for sub-tasks. We refer to the root agent as supervisor Agent. The supervisor Agent is required to perform task planning, break down the task, assign sub-tasks, and facilitate communication between specialist agents. 

As shown in Figure~\ref{fig:overview}, such a hierarchy can extend to multiple levels, with leaf agents acting as supervisors for other specialist agents. This hierarchical approach allows each leaf agent to focus on their specialized tasks, while the Supervisor Agent manages planning, delegation, and coordination. Unlike building a monolithic agent capable of solving a wide range of tasks, this approach enables the LLM behind each agent to maintain a limited context relevant to their specific role. Additionally, developing and benchmarking specialist agents becomes more manageable, and the development process can be distributed across multiple agent developers. Note that although the supervisor agent can delegate tasks to other agents, it is still the responsibility of the supervisor agent to complete the task and bring the results back to the user.

\paragraph{Inter-Agent Knowledge Exchange} Knowledge exchange between agents is a fundamental aspect of multi-agent collaboration. In this technical report, we focus on enabling the most basic form of knowledge exchange: message passing. Each agent can send direct messages to other visible agents. Within the context of hierarchical agents, an agent can send and receive messages from its supervisor and, potentially, from specialist agents. We initially support synchronized communication, where message passing temporarily blocks the execution of the sender agent until a response is received. This protocol can be extended to support asynchronous communication, allowing the sender agent to continue execution while waiting for a response.

\paragraph{Multi-Agent Evaluation} Benchmarking single AI agents is already difficult~\cite{kapoor2024ai} and increasing the number of agents to benchmark only complicates the problem. While an ideal approach is to have a human try out the multi-agent system and evaluate whether every conversation is successful or not, this kind of online evaluation is too expensive and not scalable for fast prototyping. To rapidly evaluate our multi-agent systems, we introduce assertion-based benchmarking as a way to leverage model-based evaluation and avoid dependency on collecting ground-truth conversation trajectories. For assertion-based benchmarking, we collect 90 scenarios from three different enterprise application domains. We open-source the benchmarking dataset and conversation evaluation script.\footnote{\url{https://github.com/aws-samples/multiagent-collab-scenario-benchmark}}

For enterprise applications, efficient collaboration is crucial, as many are latency-sensitive. To address this, we further optimize the multi-agent collaboration framework by introducing payload referencing and a dynamic routing mechanism. The remainder of this report provides details on methodologies and optimizations in Section~\ref{sec:approach}. Evaluation results and analysis are presented in Sections~\ref{sec:evaluation} to~\ref{sec:ablations}. Finally, we discuss limitations and future work in Sections~\ref{sec:discussion} to~\ref{sec:conclusion}.
\section{Related Work}


\paragraph{Multi-agent GenAI Systems} An agent is defined as ``an entity which is placed in an environment and
senses different parameters that are used to make a decision
based on the goal of the entity.''~\cite{kapoor2024ai}. The motivation for multi-agent systems is to have agents collaborate on a complex task that could not have been accomplished by a single agent. With the rise of LLMs, researchers have proposed leveraging the reasoning and planning capabilities of these models to develop more sophisticated multi-agent systems. Examples of such multi-agent LLM systems include MetaGPT~\cite{hong2024metagpt} and CAMEL~\cite{li2023camel}.  MetaGPT is one of the first multi-agent LLM projects that tries to mimic a software company. Developers can provide a standard operating procedure and MetaGPT tries to assign roles to various agents.
CAMEL, or Communicative Agent Framework, promotes independent collaboration between LLM agents. Its key innovation is the use of "inception prompting," a method that steers conversational agents to complete tasks. Beyond its practical applications, CAMEL also functions as a research platform. It enables the creation and analysis of conversational data, offering valuable insights into the behavior and interactions of communicative agents. Across these works, much of the emphasis is on customization and coordination, which was much less prominent in traditional multi-agent systems~\cite{singh1994multiagent,sycara1998multiagent}.

\paragraph{Multi-agent Frameworks and Platforms} CrewAI~\cite{crewai} is designed to enhance task execution by organizing agents into specialized roles, similar to team members in a crew. This approach emphasizes task decomposition, where a complex problem is broken down into smaller, manageable subtasks. Each agent is assigned a specific role based on its expertise, allowing for efficient problem-solving. AutoGen~\cite{wu2024autogen} represents a significant advancement in enabling multi-agent systems to engage in sophisticated conversations. This framework allows agents to communicate and collaborate by sharing information and refining their outputs through iterative interactions. By simulating human-like dialogues, AutoGen enables agents to negotiate, plan, and execute tasks collaboratively. 

LangGraph~\cite{langgraph} introduces an innovative framework for organizing agent interactions using directed acyclic graphs (DAGs). This structure allows for clear visualization and management of dependencies between tasks and agents. By leveraging DAGs, LangGraph optimizes the flow of information and decision-making processes among agents. This approach enhances the system’s ability to handle complex, interdependent tasks by ensuring that each agent’s actions are informed by the outcomes of preceding steps. 


\paragraph{End-to-end Agent Evaluation Frameworks}

While our work on assertion-based benchmarking is ongoing, there are other works in the literature that are similar. AgentEval~\cite{arabzadeh-etal-2024-assessing} has a multi-agent setup where there are three agents to evaluate conversation trajectories: 1) CriticAgent, 2) QuantifierAgent, 3) VerifierAgent. The CriticAgent takes the task description as input and outputs a list of criteria for task success. The QuantifierAgent then assesses whether the trajectories meet the criteria and the outputs here can be a scalar value. The VerifierAgent will finally verify that the evaluation is accurate and complete. Note that the criteria proposed by the CriticAgent is much more coarse-grained like “clarity” and “efficiency”.

AgentQuest~\cite{gioacchini-etal-2024-agentquest} proposes to measure success with a “progress rate” that is based on a set of milestones. The progress rate measures how many milestones are completed and milestones are defined to be “environment hidden states the agent needs to reach to get the final solution of the task”. AgentQuest focuses benchmarking agents for game-like datasets like ALFWorld and Sudoku, so state naturally comes with the environment when the agent plays the game. These milestones can either be externally annotated or programmatically defined within the simulation.

Most recently, ToolSandbox~\cite{lu2024toolsandbox} is the work in literature most similar to our evaluation setup. ToolSandbox also includes a user simulator and environment executor for tools. Before the simulations, they also pre-define “milestones” and “minefields” for each session. Milestones are events that must occur during the conversation and minefields are those that should not happen. These milestones and minefields are similar to our assertions that cover actions and agent behaviors. In addition, their implementation stores an “Execution Context” that contains the “world state” to mimic tasks that manipulate a resource like a database.

\section{Modeling}

\label{sec:approach}

Multi-agent collaboration enables developers to combine specialized agents to solve complex problems. Each agent can be independently developed, optimized, and configured to leverage its unique strengths. Compared to single-agent workflows, multi-agent collaboration integrates the complementary capabilities and expertise of agents with different specializations, making it highly effective for addressing complex tasks. Developers can achieve amplified capability, flexibility, and scalability by deploying a team of agents. Moreover, multi-agent solutions can have higher robustness and fault tolerance by using redundant agents in the team. For complex tasks, multi-agent solutions can improve efficiency by distributing the tasks to multiple agents and parallelizing the execution.

From a developer experience perspective, the development process is simplified by dividing functionalities among multiple agents. Developers can potentially reuse and compose existing agents for different multi-agent solutions. In some cases, cost-effective solutions can be built by utilizing low-cost orchestrating LLMs for specific agents.
In this section, we review the primary features of MAC, which include inter-agent communication, payload referencing, and dynamic agent routing. 

\subsection{Inter-Agent Communication}

\begin{figure}[t]
    \centering
    \includegraphics[width=0.9\columnwidth]{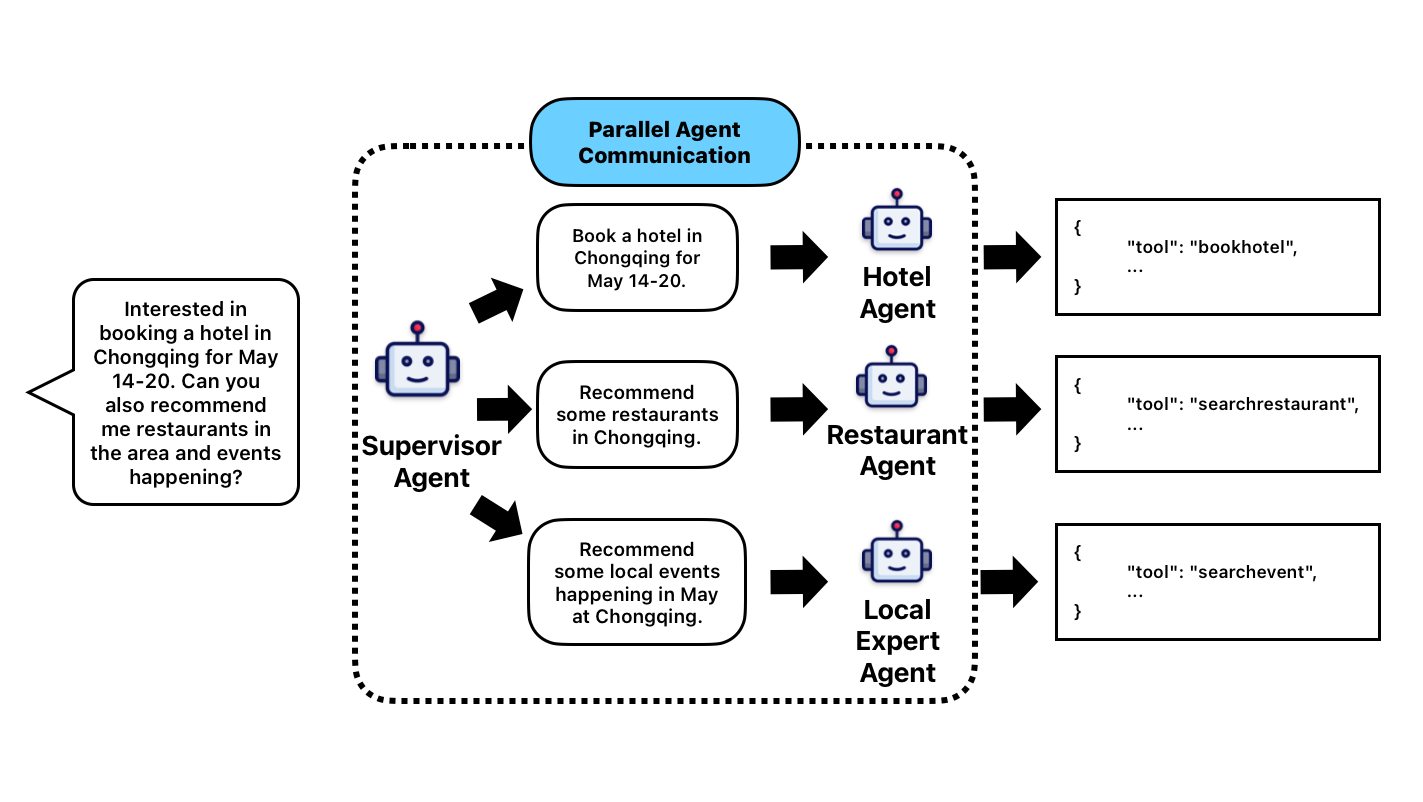}
    \caption{Example of parallel agent communication. In this example, the supervisor agent simultaneously communicates with multiple agents as the tasks can be completed independently.}
    \label{fig:parallel_comm}
\end{figure}

We model the inter-agent communication capability as a specialized tool that can be leveraged by the supervisor agent. This approach allows us to seamlessly extend the agents' communication abilities by integrating it with the existing function calling capability. The key aspects of this approach are:

\begin{enumerate}
    \item \textbf{Unified Communication Interface:} The user is treated as another agent in the system, allowing for a consistent communication interface across all interactions \textemdash \ whether it's between the user and supervisor agent, or between the supervisor agent and specialist agents.
    \item \textbf{Parallel Communication:} The supervisor agent can engage in parallel communication with multiple specialist agents simultaneously, enabling more efficient task completion through concurrent information exchange (Figure~\ref{fig:parallel_comm}).
    \item \textbf{Leveraging Existing Function Calling Capability:} By modeling communication as a tool, the supervisor agent can utilize the same underlying mechanisms for function calling to facilitate inter-agent messaging. This ensures a cohesive integration with the foundational model's existing tool-use capabilities.
\end{enumerate}

We provide the supervisor agent with a tool called \verb|send_message|, which has two parameters: \verb|recipient| and \verb|content|. This tool allows the supervisor agent to send messages to other agents. Additionally, the incoming messages from specialist agents are tagged in the following format:

\texttt{<message from="\$SOURCE\_AGENT">\newline
...\newline
</message>}

Overall, this approach to inter-agent communication helps to create a more unified and extensible multi-agent collaboration framework.

\subsection{Payload Referencing}

\begin{figure}[t]
    \centering
    \includegraphics[width=0.9\columnwidth]{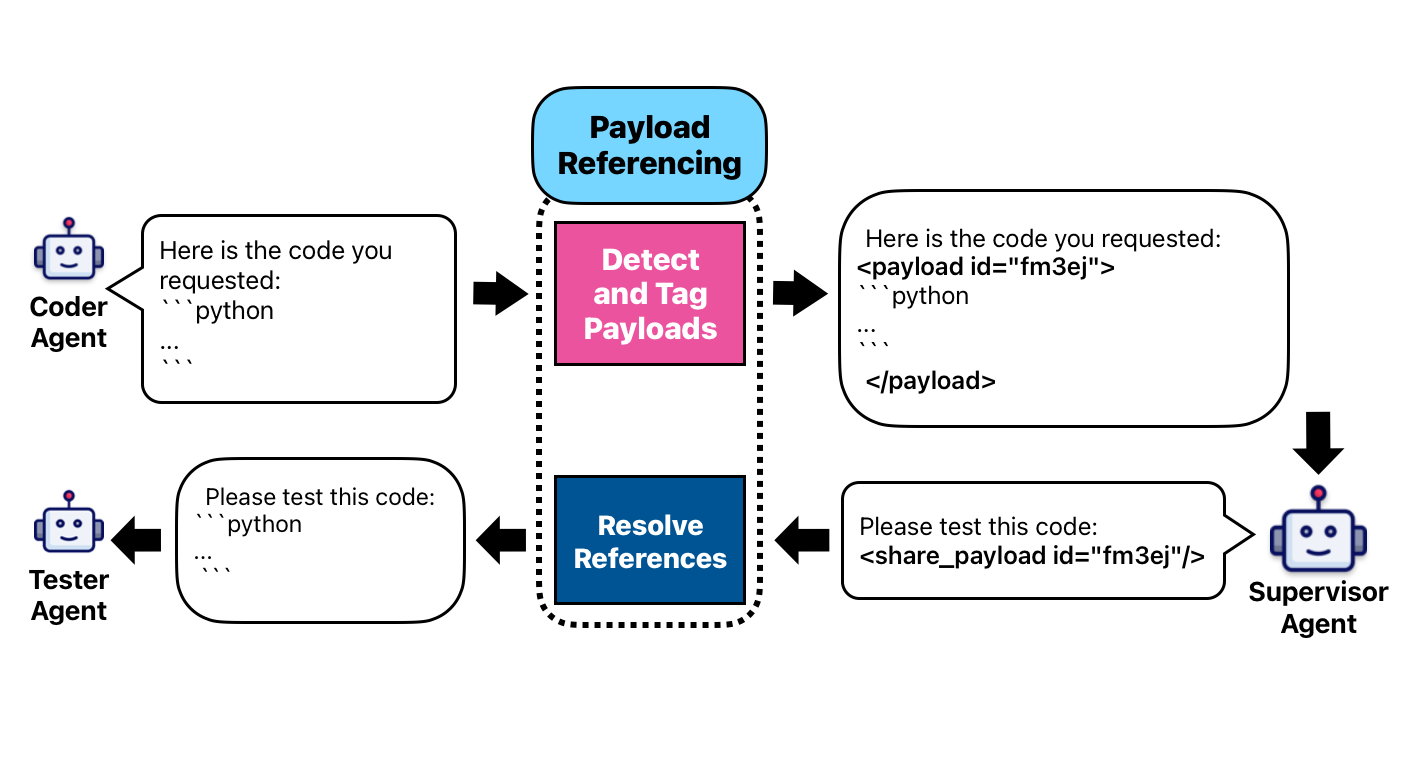}
    \caption{Example of payload referencing mechanism. In this example, the Coder agent delivers code which is then detected and tagged. The supervisor agent can then use the tag as a reference which would then be expanded to the original content for the Test agent.}
    \label{fig:payload_ref}
\end{figure}

Payload referencing is a specialized mechanism designed to handle the exchange of large content blocks, particularly code snippets (Figure~\ref{fig:payload_ref}). This mechanism aims to reduce the latency of the supervisor agent by allowing direct injection of text extracted from past multi-party communication into the message content. This is an important optimization for inter-agent communication, as the supervisor agent often needs to provide relevant context from previous interactions when communicating with specialist agents.

For example, let's say the supervisor agent (Agent A) needs to ask a specialist agent (Agent B) to perform a specific task based on the output of another specialist agent (Agent C). Instead of having Agent A regenerate the full context and details of the message from Agent C, the payload referencing mechanism allows Agent A to simply reference the relevant text from its past interactions. This can significantly reduce the number of output tokens required in the message to Agent B, leading to faster communication and reduced latency.

When a specialist agent generates a message containing structured content (e.g., code blocks), the system automatically detects these sections. For each incoming message to the supervisor agent, the detected content blocks, referred to as payloads, are assigned unique identifiers and wrapped with special tags that include these identifiers before being sent to the supervisor agent. Instead of repeatedly regenerating large static payloads in its outgoing messages to other specialist agents, the supervisor agent is instructed to reference previously detected payloads using the assigned identifiers. This allows the supervisor agent to use a simplified reference tag when sending messages. For every outgoing message from the supervisor agent, the system detects these reference tags and replaces them with the corresponding payloads before sending them to the other specialist agents. This technique enables the supervisor agent to reference payloads by generating a significantly reduced number of tokens compared to generating the entire payload itself.

Our ablation experiments with the payload referencing capability 
demonstrated a 27\% relative reduction 
in the average communication overhead per turn of the supervisor agent. This is discussed in more detail in Section \ref{sec:ablations-payload}.

\subsection{Dynamic Agent Routing}

\begin{figure}[t]
    \centering
    \includegraphics[width=0.8\columnwidth]{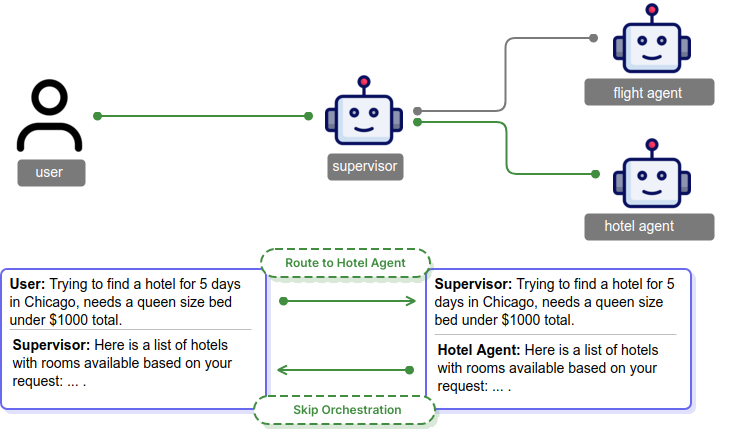}
    \caption{Dynamic agent routing, where an incoming request can be routed directly to a specialist agent, with their messages relayed back to the user.}
    \label{fig:routing}
\end{figure}

Given a complex problem, a supervisor agent often needs to communicate with multiple specialist agents across several rounds to complete the task. However, when the incoming request is simple and relevant to only a single specialized agent, triggering the full coordination capability of the supervisor agent introduces unnecessary overhead, slowing down the collaboration. In such cases, the supervisor agent only needs to route the incoming message to the appropriate specialist agent, avoiding the inefficiency of the full orchestration process.

To address this communication overhead, we introduce a dynamic agent routing mechanism that selectively bypasses the supervisor agent's orchestration when the incoming message only requires simple routing, such as the example shown in Figure~\ref{fig:routing}. The routing decision is made using a fast classifier that predicts whether the incoming message can be directly routed without additional processing. If the classifier is uncertain, it can still trigger the full orchestration process as a fallback. Note that the supervisor agent is always aware of the routing actions, and the communication between the requester and specialist agents will be available in the agent memory even the orchestration process is bypassed.

Applying dynamic agent routing can substantially improve the efficiency of multi-agent collaboration, particularly for latency-sensitive use cases. However, the success of dynamic agent routing relies on an accurate classifier capable of determining when a request requires the supervisor agent's processing with low latency. In our experiments, we demonstrate that this classification step can achieve $\ge 90\%$ accuracy with a latency of approximately 350 ms.

\section{End-to-end Automatic Evaluation}
\label{sec:evaluation}

Recent literature have noted the challenges associated with LLM agents benchmarking, which are often due to the dynamic and complex nature of the problem. The definition of success is often unclear as it can refer to either from the user perspective or from the environment/system, as users may not fully know what happens “behind-the-scenes”.  Moreover, benchmarks may not consider that user feedback can help agents achieve their goals. If the user is not properly simulated, then the evaluation undermines the agent’s capability to orchestrate tasks with a human in the loop.

Prior single-agent benchmarking is more static where user inputs and follow-up responses are pre-defined before evaluation. The gold-truth actions would be collected along with the conversations, which would then be compared to the actions generated by the agent. This static setup already has some issues because it assumes that the user goals can only be fulfilled through executing a certain set of actions. In reality, there may be multiple trajectories that enable the agent to fulfill user requests. If those trajectories are not captured in the gold-truth, then the agent is incorrectly penalized.

To generalize evaluation metrics for agents, we formally define the success of agent for a given user $u$ and scenario $s$ as $X_u^s$ , a Bernoulli random variable that represents success or failure for the user-agent session. A scenario is defined as the setting for a session which includes user goals and task domain. Since user profiles and scenarios vary, we are interested in the expected value of success for any user-agent session:

\begin{equation}
\mathbb{E}_{U,S}[X_u^s] = p_{\text{success}}    
\end{equation}

where $p_{\text{success}}$   is the success rate of an agent with a sampled user from user pool $U$ and scenario sampled from collection $S$.  The objective of benchmarking is to approximate $p_{\text{success}}$.

We can easily extend this formulation to multi-agent systems. However, with more agents added, the complexity grows and $p_{\text{success}}$ is more difficult to approximate. Several, different trajectories could be considered correct and sometimes no actions need to be executed to achieve user goals.  We propose \textbf{assertion-based benchmarking} as a way to better approximate likelihood of agent success.

\begin{figure}[t]
    \centering
    \includegraphics[width=0.9\columnwidth]{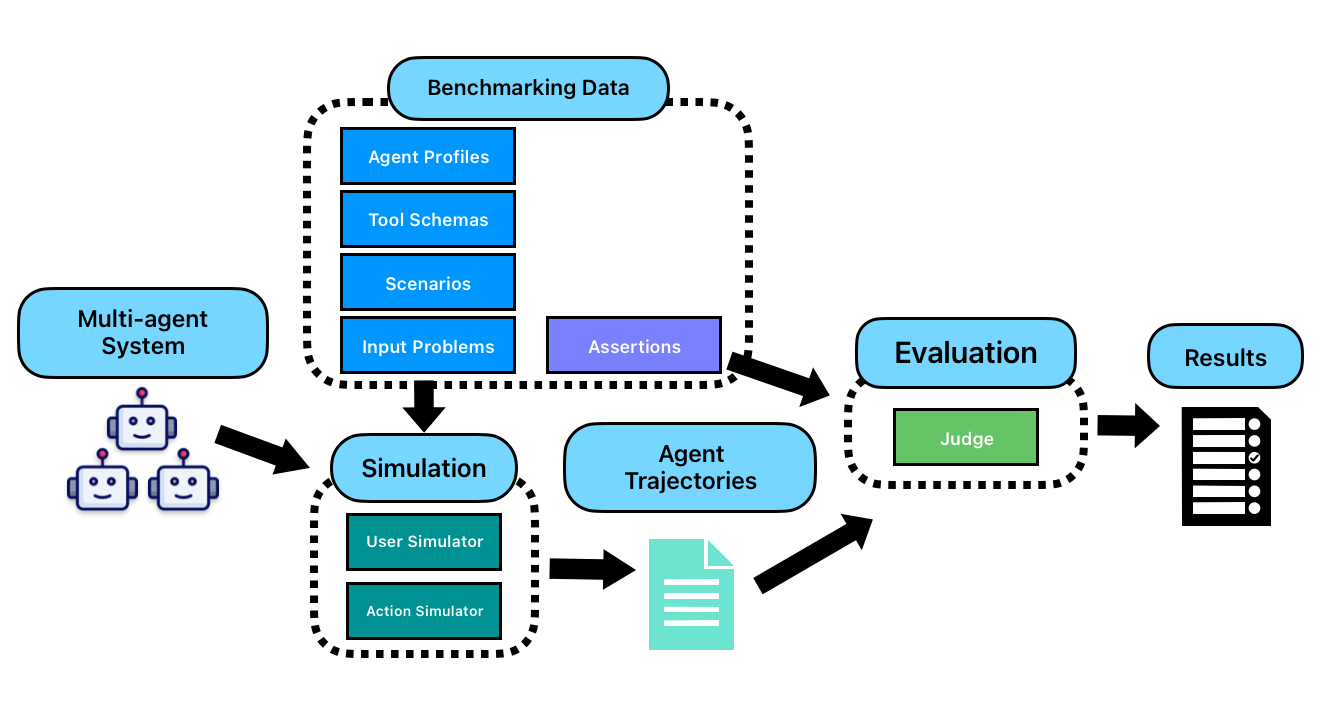}
    \caption{Overview of end-to-end assertion-based benchmarking with scenarios and assertions}
    \label{fig:benchmarking}
\end{figure}

The assertion-based evaluation framework relies on three components: 1) benchmarking data collection, 2) environment simulators, 3) automatic assertion judge (Figure~\ref{fig:benchmarking}). First, a set of scenarios need to be collected according to pre-defined agent profiles and tool schemas for a particular domain. With each scenario, a list of assertions needs to be included. These assertions are statements that must hold true for a conversation to be considered successful. This is similar to debugging software with test cases. Second, the user simulator is initialized from the scenario and input problem and begins to interact with the multi-agent system. An action simulator is also used to simulate the tool calls from the multi-agent system. The trajectories are recorded for further evaluation. Finally, a judge determines the validity of each assertion. Based on the number of correct assertions, goal success rates can be computed to help measure the success of the multi-agent system. In the following subsections, we cover each component of the framework in depth.

\subsection{Benchmarking Data}

There are three artifacts that need to be collected for each scenario. First, the scenario description itself must list the user goals and any background information. The information in the scenario description is critical because the user simulator will be grounded on the description. Then, the second artifact is the input problem, which is the first turn of the conversation from the user. The input problem will begin the benchmarking simulation. 

The final artifact is a list of assertions that need to be satisfied during the simulation. Assertions are categorized two types: user-side and system-side. User-side assertions cover behaviors of the multi-agent system that can be observed by the user. On the other hand, system-side assertions cover behaviors of the multi-agent system that cannot be observed by the user.  These assertions may include tool calling correctness, parameter correctness, inter-agent behavior, or rule compliance. Appendix~\ref{app:benchmark_artifacts} shows example artifacts from our publicly released benchmarking data collection.

\subsection{Simulation}

For each session, we feed the scenario description and the input problem to a user simulator. The user simulator is a LLM that is prompted to follow the scenario description. The simulation begins with the input problem as the first turn of the session. The input problem is delivered to the supervisor agent who then begins to work the specialist agents. If the specialist agents need to invoke an action, the function calls are passed to the action simulator, which is a LLM grounded on the provided tool schemas.  The action simulator also has access to past tool invocations so that it can generate results that is aligned to past observations. After the action simulator returns the simulated action, the specialist agents continue to carry out the task.

During the simulation, the user simulator will continue to interact with the supervisor agents. The user simulator can either help clarify any questions from the supervisor agent or keep sending new task requests.  Any requests or answers given by the user simulator should be aligned with the information in the scenario. Once the user simulator determines that all the goals in the scenario are met, the user simulator generates a \texttt{</stop>} token to end the simulation. Otherwise, we set a maximum number of user simulation turns to 5 to prevent simulations that fail to end.

\subsection{Assertion Judgements}

After simulations are completed, we pass the trajectories to a LLM judge to help automate the assertion evaluation. Along with the simulation trajectories, we also pass the scenario and the assertions to the judge. The judge returns whether each assertion is True or False, and includes the reason for their judgement. The reason often exhibits evidence from the conversation, which can easily show why the assertion has succeeded or failed. This is important as multi-agent conversations can be very lengthy and difficult for people to pinpoint the causes of failures.

\subsection{Metrics}

\begin{table}[t]
    \caption{Definitions of Success Metrics}
    \label{tab:gsr_define}
    \centering
    \footnotesize
    \begin{tabular}{p{2cm}p{5cm}p{5cm}}
    \toprule
    Metric  & Definition & Implementation  \\
    \midrule
    Overall GSR & Overall goal success rate covering both the user-side and the system-side & Use LLM to judge user-side and system-side assertions. For a conversation, score is 1 if all assertions are True; else 0. \\
    \midrule
    Supervisor GSR & Goal success rate of the supervisor agent without any dependence on sub-agent and tool behavior & If overall GSR is 1 or supervisor agent is reliable (see reliability metrics), then score for the conversation is 1; else 0. \\
    \midrule
    User-side GSR & Goal success rate in the perspective of the user & Use LLM to judge user-side assertions. For a conversation, score is 1 if all user-side assertions are True; else 0. \\
    \midrule
    System-side GSR & Goal success rate in the perspective of the system developers & Use LLM to judge system-side assertions. For a conversation, score is 1 if all system-side assertions are True; else 0. \\
    \bottomrule
    \end{tabular}
\end{table}

A conversation is considered overall successful if all assertions, both user-side and system-side, are True. We then measure \textbf{Goal Success Rate (GSR)} as the percentage of conversations that have all assertions evaluated as True. This overall GSR is our primary measure of success. We also compute User GSR and System GSR, which are the variants of GSR where the assertions being evaluated are limited to one type (either user-side or system-side). These metrics are useful to understand whether multi-agent systems are failing from the perspective of the user or the system (Table~\ref{tab:gsr_define}).

\begin{table}[t]
    \caption{Definitions of Latency Metrics}
    \label{tab:latency_define}
    \centering
    \footnotesize
    \begin{tabular}{p{3cm}p{9cm}}
    \toprule
    Metric  & Definition  \\
    \midrule
    Avg. communication overhead per turn & Average number of seconds that the supervisor agent spends communicating with other agents before getting back to the user. This time does not take into account the duration of agents other than the supervisor agent. \\
    \midrule
    Avg. latency per communication & Average number of seconds that the supervisor agent spends to deliver each message to communicate with other agents. \\
    \midrule
    Avg. user-perceived turn latency per session & Average number of seconds it takes for the supervisor agent to get back to the user. This time does take into account the duration of all agents in the system. \\
    \midrule
    Avg. communications per session & Average count of messages sent by the supervisor agent over the entire session. \\
    \midrule
    Avg. output tokens per communication & Average number of total output tokens from the supervisor agent for each message. \\
    \bottomrule
    \end{tabular}
\end{table}

Beyond goal success, latency is a critical metric in multi-agent systems. As these systems involve multiple agents interacting and collaborating to perform complex tasks, the time delay between agent communications and actions can significantly impact user experience. Minimizing latency is crucial for ensuring the system can operate efficiently, especially in time-sensitive applications. Table~\ref{tab:latency_define} lists the various latency metrics that are included in this report.


\begin{table}[t]
    \caption{Definitions of Routing Metrics}
    \label{tab:routing_define}
    \centering
    \footnotesize
    \begin{tabular}{p{4cm}p{8cm}}
    \toprule
    Metric  & Definition  \\
    \midrule
    Classification Accuracy & Accuracy of routing decisions, calculated based on human annotated ground-truth labels. \\
    \midrule
    False Agent Switch Rate & Ratio of routing decisions causing the handling agent to be switched to a wrong agent \\
    \midrule
    Turn-level Routing Overhead (ms) & Time taken in the primary agent (including routing classification and orchestration) within a user turn. \\
    \midrule
    Classification Latency (ms) & Time taken for classifying a single routing decision. \\
    \bottomrule
    \end{tabular}
\end{table}

Lastly, we also included metrics to evaluate routing mode. For routing mode, we are concerned with routing classification accuracy, false agent switch rates, turn-level routing overhead, and routing classification latency (Table~\ref{tab:routing_define}).
\section{Experimental Results}
\label{sec:experiments}

To understand how MAC performs for enterprise usage, we choose three business domains to experiment with. For each domain, we set up a set of agents and tools. Then, we manually collect benchmarking data for each domain. In this report, we show experiments on thirty scenarios from each domain. This dataset is also publicly released for others to benchmark their own multi-agent systems (Section~\ref{sec:intro}). The three domains are as follows:
\begin{enumerate}
\item Travel planning: agents help user plan a trip, which includes booking flights, booking hotels, finding local events, getting information about the weather, etc.
\item Mortgage financing: agents help user with mortgage issues, e.g., submitting loan applications, querying information about properties, retrieving banking information, etc.
\item Software development: agents help user design, implement, test, review, and/or deploy code.
\end{enumerate}

The first two domains are conversational, whereas the third domain is more about automation and requires minimal interaction with user. For the agent profile design, we have made sure to cover the following conditions to showcase adoption of multi-agent collaboration for various developer setups:

\begin{enumerate}
\item Supervisor agents with and without tools: The supervisor agent for Mortgage has access to MortgageLoans tools, whereas the supervisor agents for other domains do not have tools.
\item Specialist agents with and without tools: Many specialist agents have their own set of tools. However, some specialist agents in Software do not have tools as they should have the capability to complete the tasks without them.
\item Agent hierarchy with depth more than one: In Software, the supervisor agent can call on Deploy agent, which can then call on Infrastructure agent and Application agent.
\end{enumerate}

\begin{table}[t]
    \caption{Dataset statistics, including average number of goals per scenario, average number of assertions per scenario, total number of action groups, and total number of APIs/tools.}
    \label{tab:dataset_stats}
    \centering
    \footnotesize
    \begin{tabular}{lrrrr}
    \toprule
     Dataset & Avg. Goals per Scenario & Avg. Assertions per Scenario & Action Groups & APIs/Tools \\
     \midrule
     Travel & 2.13 & 4.55 & 11 & 52 \\
     Mortgage & 1.91 & 3.99 & 10 & 35 \\
     Software & 1.69 & 7.47 & 2 & 4 \\
     \bottomrule
     \end{tabular}
\end{table}

Table~\ref{tab:dataset_stats} shows an overview of the statistics for the three datasets. Action groups refer to group of tools (e.g. ``BookFlight'' action group is a set of tools for searching flights, booking flights, getting flight details, etc.). Travel and Mortgage each have more than thirty tools, whereas Software has only two tools. Software tends to have more assertions on average than Travel and Mortgage. Appendix~\ref{app:agent_profiles} shows the agents for each domain and their associated action groups, which are the set of tools that the agents have access to.

\subsection{Coordination Mode Experiments}

\begin{table}[t]
    \caption{End-to-end evaluation of Coordination Mode}
    \label{tab:coord_results}
    \centering
    \footnotesize
    \begin{tabular}{llrr}
    \toprule
    Setting & Dataset & Overall GSR & Supervisor GSR \\
    \midrule 
    \multirow{3}{*}{\shortstack[l]{Supervisor: Sonnet 3.5 (20241022) \\ Specialists: Sonnet 3.5 (20241022)}} & Travel & 0.90 & 0.90 \\
    & Mortgage & 0.90 & 0.93 \\
    & Software & 0.90 & 1.00 \\
    \midrule 
    \multirow{3}{*}{\shortstack[l]{Supervisor: Sonnet 3.5 (20241022) \\ Specialists: Sonnet 3.0}} & Travel & 0.87 & 0.90 \\
    & Mortgage & 0.90 & 0.97 \\
    & Software & 0.77 & 0.90 \\
    \midrule
    \multirow{3}{*}{\shortstack[l]{Supervisor: Sonnet 3.5 (20241022) \\ Specialists: Haiku 3.5}} & Travel & 0.80 & 0.87 \\
    & Mortgage & 0.83 & 0.90 \\
    & Software & 0.87 & 0.93 \\
    \midrule
    \multirow{3}{*}{Single-agent: Sonnet 3.5 (20241022)} & Travel & 0.60 & -- \\
    & Mortgage & 0.80 & -- \\
    & Software & 0.53 & -- \\
    \bottomrule
    \end{tabular}
\end{table}

\begin{table}[t]
    \caption{Latency Performance of Coordination Mode}
    \label{tab:coord_latency}
    \centering
    \footnotesize
    \begin{tabular}{llrr}
    \toprule
    Setting & Dataset & \shortstack[r]{Avg. communication \\overhead per turn (s)} & \shortstack[r]{Avg. user-perceived \\turn latency \\per session (s)} \\
    \midrule 
    \multirow{3}{*}{\shortstack[l]{Supervisor: Sonnet 3.5 (20241022) \\ Specialists: Sonnet 3.5 (20241022)}} & Travel & 13.75 & 31.46 \\
    & Mortgage & 13.39 & 24.42 \\
    & Software & 35.44 & 168.73 \\
    \midrule 
    \multirow{3}{*}{\shortstack[l]{Supervisor: Sonnet 3.5 (20241022) \\ Specialists: Sonnet 3.0}} & Travel & 15.43 & 42.12 \\
    & Mortgage & 15.97 & 29.90 \\
    & Software & 53.48 & 137.31 \\
    \midrule
    \multirow{3}{*}{\shortstack[l]{Supervisor: Sonnet 3.5 (20241022) \\ Specialists: Haiku 3.5}} & Travel & 12.95 & 23.98 \\
    & Mortgage & 11.03 & 18.13 \\
    & Software & 36.65 & 125.31 \\
    \midrule
    \multirow{3}{*}{Single-agent: Sonnet 3.5 (20241022)} & Travel & -- & 14.12 \\
    & Mortgage & -- & 9.12 \\
    & Software & -- & 52.61 \\
    \bottomrule
    \end{tabular}
\end{table}

We report the goal success metrics for coordination mode across a variety of settings in Table~\ref{tab:coord_results}. For evaluating the assertions, we leverage OpenAI’s GPT-4o model~\cite{gpt4o} for providing LLM-based judgements. The results demonstrate that multi-agent collaboration achieves the best overall GSR of 90\% across the three evaluated domains when the Claude 3.5 Sonnet (20241022) model~\cite{claude35} is utilized as both the supervisor agent and the specialist agents. 

We further examine the impact of using different agent models. Switching the specialist agent model from Claude 3.5 Sonnet  to Claude 3.0 Sonnet shows a significant regression in the performance for the Software development domain. This is also apparent when the specialist agent is switched to Claude 3.5 Haiku, as the performance in the Software domain is mostly recovered with this newer generation of models.

We also compare the performance of a single agent with the multi-agent collaboration approach. In this experiment, a single agent is given the tools from all specialist agents combined and is responsible for assisting the user. We reuse the assertions collected for the multi-agent setup but replace any mentions of specialist agents with supervisor agent (e.g. ``flight agent books tickets'' becomes ``travel agent books tickets''). For any assertions about inter-agent behaviors, we would replace mentions of the primary agent with ``user'' and mentions of the specialist agent with the supervisor agent (e.g. ``code agent implements code and delivers back to software agent'' becomes ``software agent implements code and delivers back to user'').  

In the single-agent setting, we observe an absolute regression of up to 37\%. MAC allows each specialist agent to be provided with instructions for the specific subset of tasks it is supposed to handle. This specialized task assignment may not be achievable by a single agent, which may struggle to manage the multitude of instructions required to complete the complex tasks. In the single-agent trajectories, we observe more hallucination in tool parameters and incorrect tool choice.

We report the latency metrics for our experiments in Table~\ref{tab:coord_latency}. When using Claude 3.5 Sonnet (20241022) for both supervisor agent and specialist agents, the average communication overhead per turn ranges from 13.39s to 35.44s, with the Software domain showing significantly higher overhead due to its complexity. The Software domain consistently demonstrates higher latency metrics across all settings, with user-perceived turn latency reaching 168.73s compared to 31.46s for Travel and 24.42s for Mortgage domains. While the single-agent setting shows lower user-perceived latencies, this comes at the cost of reduced goal success rates as shown in the previous results. See Appendix~\ref{app:full} for full results on coordination mode.

We also performed ablation experiments with the payload referencing capability of the supervisor agent, which is discussed in more detail in Section \ref{sec:ablations-payload}.
\subsection{Routing Mode Experiments}


To benchmark dynamic agent routing, we curated two additional datasets for evaluation from our original benchmarking data: 1) Mortgage Routing (3 agent layers) and 2) Travel Routing (2 agent layers), and manually annotated around 100 routing classification labels for each datasets. Intrinsic evaluation shows that the LLM-based routing solution with Claude 3.0 Haiku achieves more than 90\% routing classification accuracy and less than 3\% false agent switching rate (Table~\ref{tab:routing_results}). Average latency of routing classification is about 350 ms. With end-to-end evaluation, we observe turn-level routing overhead (time taken by the supervisor agents) in 600 ms to 800ms range (Table~\ref{tab:routing_results2}). 

\begin{table}[t]
    \caption{Routing Mode Performance}
    \label{tab:routing_results}
    \centering
    \footnotesize
    \begin{tabular}{lrrrrr}
    \toprule
    Dataset & \shortstack[r]{Classification \\accuracy} & \shortstack[r]{False \\ switch \\ rate} & \shortstack[r]{Turn-level \\ routing \\overhead (ms)} & \shortstack[r]{Classification \\ latency (ms)} \\
    \midrule 
    Mortgage Routing (First Layer) & 0.92 & 0.00 & 630 & 344 \\
    Mortgage Routing (Second Layer) & 0.92 & 0.00 & 630 & 378 \\
    Travel Routing & 0.90 & 0.03 & 750 & 360 \\
    \bottomrule
    \end{tabular}
\end{table}

\begin{table}[t]
    \caption{End-to-end evaluation of Routing Mode}
    \label{tab:routing_results2}
    \centering
    \footnotesize
    \begin{tabular}{lrrr}
    \toprule
    Domain    & Overall GSR & Supervisor GSR & \shortstack[r]{Avg. Routing Overhead \\ per Turn (ms)} \\
    \midrule
    Travel & 0.85 & 1.00 & 751 \\
    Mortgage & 0.95 & 1.00 & 627 \\
    \bottomrule
    \end{tabular}
\end{table}
\subsection{Agreement with Human Evaluation}

Throughout this report, we have used an LLM to provide automatic judgements on assertions. This has helped scale our experiments so that we can quickly prototype and develop improved multi-agent systems. What if we ask a human to judge conversation success purely from their own perspective without any reference to assertions? To understand how well assertion-based evaluation compares with human evaluation, we deliver a batch of ninety trajectories to human annotators during one of our milestone checkpoints. For each conversation, we ask three annotators to provide binary judgements on a different set of guidelines. During this human evaluation, they are never shown assertions or LLM judgements and only instructed to determine success and efficiency from their own judgements. Table~\ref{tab:human_instructions} show the instructions given to human annotators for each success metric.

After the human annotators finish their evaluation, the aggregated majority judgement is used to compare against LLM evaluations. Note that for this milestone, we use Claude 3.5 Sonnet (20240620) for supervisor agent and Sonnet 3.0 for specialist agents.  See Appendix~\ref{app:human_eval} for the automatic and human evaluation results for this set of experiments.  We then compute agreement between human and LLM judges. Since the judgements for automatic and human evaluation are binary, we use agreement ratio to measure alignment between human and automatic evaluation. The agreement ratio is the number of conversations with matching judgements (either both 1 or both 0) over the total number of conversations evaluated. 

\begin{table}[t]
    \caption{The instructions given to annotators when performing human evaluation of conversation trajectories. Note that humans do not judge assertions and only determine success purely from their own judgements.}
    \label{tab:human_instructions}
    \centering
    \footnotesize
    \begin{tabular}{p{2cm}p{10cm}}
    \toprule
    Metric & Human annotator instructions \\
    \midrule
    User GSR & Evaluate from user’s perspective, whether the conversation with the primary agent meets the user's goals and successfully address user's requests or not. \\
    System GSR & From the perspective of the environment, evaluate whether the actual impacts of all agent actions accurately address and resolve all of the user's stated tasks, requests, and expectations within that specific environment or context.  \\
    Supervisor GSR & Evaluate from user's perspective whether the primary agent has tried its best to help the User, regardless of whether task was completed or the agent actions correctly impacted the environment. \\
    \bottomrule
    \end{tabular}
\end{table}

\begin{table}[t]
    \caption{The agreement ratios between human annotators and assertion-based benchmarking on success metrics across 90 conversation trajectories.}
    \label{tab:human_agreement}
    \centering
    \footnotesize
    \begin{tabular}{lrrrr}
    \toprule
    Dataset & Overall GSR & Supervisor GSR & User-side GSR & System-side GSR \\
    \midrule 
    Travel & 0.93 & 0.87 & 0.93 & 0.93 \\
    Mortgage & 0.87 & 1.00 & 0.97 & 0.87 \\
    Software & 0.97 & 0.77 & 1.00 & 0.90 \\
    \bottomrule
    \end{tabular}
\end{table}

For success metrics, the agreement is generally above 85\% (Table~\ref{tab:human_agreement}). The only exception is on primary agent success for software, where the agreement is 77\%. Here, the automatic evaluation has overall GSR as 90\% but the human annotators only believe the supervisor agent is successful in 87\% of the conversations. On the trajectories where the humans disagree with LLM judgements, we observe a mix of mistakes from both ends. For some trajectories, the human evaluation would mark that supervisor agent has not make a mistake but LLM judge has captured when the supervisor agent repeatedly wanting to follow-up with the user rather than delegating tasks to sub-agents. There is also the reversed case where the LLM judge does not observe any issues with the supervisor agent but the human annotators detect that the supervisor agent neglects to conduct thorough code review with the Review agent.


\section{Communication Mechanism Ablations}
\label{sec:ablations}

Section~\ref{sec:experiments} shows the main results of our approach with different models. In this section, we provide additional ablations to quantify the impact of MAC communication mechanisms. This includes experiments with and without payload referencing, as well as comparison with open-source frameworks.  The additional experiments provide more justification on the utility of MAC for enterprise applications.

\subsection{Impact of Payload Referencing}
\label{sec:ablations-payload}

\begin{table}[t]
    \caption{Impact of payload referencing on GSR and latency for Software}
    \label{tab:payload_ref}
    \centering
    \footnotesize
    \begin{tabular}{lrrrr}
    \toprule
     & \shortstack[r]{Overall \\ GSR} & \shortstack[r]{Avg. \\ communication \\ overhead per \\ turn (s)} & \shortstack[r]{Avg. \\ user-perceived \\ turn latency \\ per session (s)} & \shortstack[r]{Avg. \\ output \\ tokens per \\ communication} \\
    \midrule 
    With Payload Ref. & 0.90 & 35.44 & 168.73 & 373.77 \\
    Without Payload Ref. & 0.73 & 48.78 & 159.75 & 539.21 \\
    \bottomrule
    \end{tabular}
\end{table}

In Table~\ref{tab:payload_ref}, we report the results of an ablation experiment with Payload Referencing for Software domain with Claude 3.5 Sonnet (20241022) as the supervisor agent as well as specialist agents. This experiment reveals that payload referencing significantly improves both the efficiency as well as effectiveness of multi-agent collaboration, particularly in the code-heavy Software domain. The impact is most pronounced here as large code blocks are frequently exchanged between agents.

Enabling payload referencing results in a 23\% relative improvement in overall GSR as well as a 27\% relative reduction in the average communication overhead per turn. The latter can be attributed to a 30\% relative reduction in the average output tokens per communication of the supervisor agent, as it is able to leverage the payload referencing mechanism to more efficiently reduce the number of generated tokens required for sharing code blocks, while improving the goal success rate.

When enabling payload referencing, we also observe an increase in the average user-perceived turn latency. As this also includes the latency of the specialist agents, it suggests that the specialist agents operating for a smaller number of turns may be detrimental to the overall goal success.

In conclusion, payload referencing proves particularly valuable in domains such as Software, as it:

\begin{itemize}
    \item Enables precise referencing of payloads without expensive regeneration of tokens
    \item Reduces the likelihood of payload corruption during agent-to-agent transmission
    \item Maintains formatting and structure of the payload across agent communications
    \item Facilitates more efficient parallel inter-agent communication
\end{itemize}

Our experiments suggest that payload referencing is a crucial optimization for multi-agent systems, particularly in domains involving structured content exchange. The mechanism not only improves system performance metrics but also enhances the quality and reliability of agent interactions.

\subsection{Comparison with Task Automation Framework}
\begin{table}[t]
    \caption{Comparison with a widely adopted open-source framework for task automation (OSF). Note that the version of Sonnet 3.5 used below refers to an older version.}
    \label{tab:osf_results}
    \centering
    \footnotesize
    \begin{tabular}{llrr}
    \toprule
    Setting & Dataset & Overall GSR & Supervisor GSR \\
    \midrule 
    \multirow{3}{*}{\shortstack[l]{[MAC Coordination Mode] \\ Supervisor: Sonnet 3.5 (20240620) \\ Specialists: Sonnet 3.0}} & Travel & 0.87 & 0.90 \\
    & Mortgage & 0.90 & 0.97 \\
    & Software & 0.77 & 0.90 \\
    \midrule 
    \multirow{3}{*}{\shortstack[l]{[OSF] \\ Supervisor: Sonnet 3.5 (20240620) \\ Specialists: Sonnet 3.0}} 
    & Travel & 0.50 & 0.80 \\
    & Mortgage & 0.63 & 0.87 \\
    & Software & 0.40 & 0.63 \\
    \midrule
    \multirow{3}{*}{\shortstack[l]{[OSF] \\ Supervisor: GPT-4o (mini) \\ Specialists: GPT-4o (mini)}} 
    & Travel & 0.43 & 0.63 \\
    & Mortgage & 0.40 & 0.67 \\ 
    & Software & 0.33 & 0.77 \\
    \bottomrule
    \end{tabular}
\end{table}

We also conducted a comparative evaluation of our implementation against a widely adopted open-source framework (OSF) for task automation. The OSF implements multi-agent collaboration as a sequential task-automation workflow, wherein the supervisor agent receives a task from the user and is responsible for breaking it up into one-time sub-tasks for its sub-agents. However, as our evaluation is primarily based on more conversational back-and-forth between the specialist agents, the one-time task assignment may not be sufficient. Therefore, we also expanded the OSF to support multiple sub-agent interactions. The OSF also enables tool-use for its agents using ReAct-style prompting~\cite{yao2023react}, as opposed to native function-calling capability of our implementation. 

In Table~\ref{tab:osf_results}, we present a comparative analysis with the OSF. Due to the prompting style of the agents in the OSF, they tend to be highly verbose and utilize a much higher number of tokens. To constrain the experiment to a reasonable budget, we employ Claude 3.5 Sonnet (20240620) as the supervisor agent and Claude 3.0 Sonnet as the specialist agents. Since the original prompts in the OSF may have been optimized for GPT-4o (mini), as suggested by the defaults in the code implementation, we also report the performance of GPT-4o (mini) with the OSF in the conversational setting.
\section{Discussion}
\label{sec:discussion}

The results presented in this technical report demonstrate the effectiveness of MAC in enabling coordinated problem-solving through multiple specialized agents. Across the evaluated domains of Travel, Mortgage, and Software Development, the multi-agent collaboration approach achieved impressive goal success rates, with the overall GSR reaching as high as 90\% when utilizing the Claude 3.5 Sonnet (20241022) models for both the supervisor agent and specialist agents.
This strong performance is particularly notable when compared to the single-agent setting, where a significant regression of up to 37\% was observed in the Software Development domain. The ability of the multi-agent collaboration to leverage the specialized expertise of individual agents appears to be a key factor in this improved effectiveness.

Further analysis of the results highlights the importance of optimizing the inter-agent communication mechanisms. The impact of the payload referencing feature is most pronounced in the Software Development domain, where it led to a 23\% relative improvement in overall GSR as well as a 27\% relative reduction in the average communication overhead per turn. By allowing the supervisor agent to more efficiently reference and share large content blocks, such as code snippets, this specialized mechanism enhances the reliability and speed of the multi-agent interactions.
However, the results also reveal some limitations of the current system. The higher latency observed in the more complex Software Development scenarios suggests that further optimizations may be needed to reduce the overhead of multi-agent coordination, particularly in time-sensitive applications. 

\section{Conclusion}
\label{sec:conclusion}

This paper presents a comprehensive evaluation of the multi-agent collaboration framework, demonstrating its effectiveness in enabling coordinated problem-solving across a variety of enterprise domains. The results highlight the system's ability to leverage specialized agents to tackle complex tasks, achieving impressive goal success rates of up to 90\% in the evaluated scenarios. A key aspect of MAC is the focus on optimizing the inter-agent communication mechanisms. The introduction of the payload referencing feature, for example, was shown to provide significant benefits, particularly in code-heavy tasks, by allowing the supervisor agent to more efficiently share and reference large content blocks. This optimization  has led to remarkable improvements in both the overall goal success rate and the communication efficiency of the system.

Furthermore, the evaluation framework that is employed in this study, which combines assertion-based benchmarking with automated LLM-based judgements, has proven to be a reliable and scalable approach for assessing the performance of multi-agent collaborative systems. During development, we observe high agreement rates on goal success between human and automated evaluation. This validates our evaluation framework, which can enable faster prototyping and development of MAC.

One key focus for future work will be on further reducing the latency observed in more complex scenarios, such as those in the Software Development domain. Exploring additional optimizations to the inter-agent coordination mechanisms may help to address this challenge and ensure the system can operate efficiently, even in time-sensitive applications. Additionally, expanding the detection and handling of different forms of static long-form payloads could lead to additional performance gains. Investigating automatic prompt optimization techniques may also prove valuable in enhancing the collaboration capabilities of the individual agents.

Finally, automating the dataset curation process for the benchmarking framework could enable more scalable and iterative development of the multi-agent system, allowing for faster prototyping and deployment of improvements. By building on the solid foundations laid out in this technical report, the future work in these areas promises to further strengthen MAC and its ability to tackle increasingly complex, real-world challenges.

\begin{ack}
First, we want to thank the AWS Bedrock Agents Science team for their help in the project, notably Tamer Alkhouli, Renato Negrinho, Veera Raghavendra Elluru, and Yassine Benajiba. We thank Claudia Zaghi and Nina Rondoni for their  substantial support in data collection. We thank the AWS Bedrock Product team and solution architects for their crucial feedback on enterprise needs and usecases. Finally, we want to acknowledge the AWS Bedrock Engineering team for diligently deploying the multi-agents collaboration service and delivering our work on enterprise applications to AWS customers. 
\end{ack}

\bibliographystyle{plain}
\bibliography{refs}

\newpage
\appendix
\section*{Appendix}
\section{Benchmarking Data Artifacts}
\label{app:benchmark_artifacts}

\begin{table}[h!]
    \caption{Example artifacts from benchmarking data collection}
    \label{tab:benchmark_artifacts}
    \centering
    \footnotesize
    \begin{tabular}{p{2cm}p{10cm}}
        \toprule
        Artifact & Example \\
        \midrule
        Scenario & Goals: \newline
\texttt{* User needs to book a ticket for a round-trip economy flight from DEN to RST, departing on June 23, and returning on June 30.\newline
* User needs to book a room in Rochester, Minnesota from June 23 to June 30.\newline
* User needs to obtain total estimated cost of flight, hotel, food, and local transportation for their 7-day trip, given their \$2,500 weekly budget.}\newline \newline
Background: \newline
\texttt{* User's full name is Gregory James Anderson.\newline
* Gregory is 36 years old.\newline
* Gregory resides in Denver, Colorado.\newline
* Gregory's preferred class for flights is economy.}  \\
\midrule
Input Problem & \texttt{Please book a ticket for a round-trip economy flight from DEN to RST, departing on June 23.}\\
\midrule
Assertions & User-side assertions:\newline
\texttt{* User is informed that a ticket for an economy flight from DEN to RST departing on June 23 have been booked.\newline
* User is informed that a ticket for an economy flight from RST to DEN have been booked. The flight from DEN to RST departs on June 30.\newline
* User is informed that a room in a hotel in Rochester, Minnesota from June 23 to June 30 has been booked\newline
* User is informed of the total estimated cost including flight, hotel, food, and local transportation for their 7-day trip from Denver to Minnesota, given their \$2,500 weekly budget.}\newline \newline System-side assertions:\newline
\texttt{* book\_flight is executed to book two tickets for a round-trip economy flight from DEN to RST on June 23\newline
* book\_hotel is executed to book a hotel in Minnesota from June 23 to June 30\newline
* search\_flight is executed before book\_flight to provide the user with flight options before performing the booking\newline
* search\_hotel is executed before book\_hotel to provide the user with options before booking\newline
* calculate is executed to get the total estimated cost for flight, hotel, food, and local transportation for his 7-day trip from Denver to Minnesota for one person, given their \$2,500 weekly budget} \\
        \bottomrule
    \end{tabular}
\end{table}

\section{Agent Profiles}
\label{app:agent_profiles}

\begin{table}[H]
    \caption{List of agents for each benchmarking domain. Supervisor agents are \textbf{bolded}. We also list their action groups, which are sets of tools that the agents have access to.}
    \label{tab:agent_profiles}
    \centering
    \footnotesize
    \begin{tabular}{p{3cm}p{3cm}p{6cm}}
    \toprule
    Domain/Usecase & Agent Name & Action Groups \\
    \midrule
    \multirow{10}{*}{Travel Planning} & \textbf{Travel agent} & \\
    & Weather agent & Weather \\
    & Location search agent & LocationService \\
    & Car rental agent & CarRental \\
    & Flight agent & BookFlight \\
    & Hotel agent & BookHotel \\
    & Travel budget agent & Calculator \\
    & Restaurant agent & RestaurantSearch,  FoodDelivery \\
    & Local expert agent & Eventbrite, NewsSeartch \\
    & Airbnb agent & BookAirbnb \\
    \midrule 
    \multirow{6}{*}{Mortgage Financing} & \textbf{Mortgage agent} & MortgageLoans \\
    & Property agent & LocationService, RealEstateManagement \\
    & Credit agent & Banking, CreditReport \\
    & Income agent & HRPayrollBenefits, Calculator \\
    & Payment agent & Calculator \\
    & Closing agent & Calculator, RealEstateManagement \\
    \midrule
    \multirow{8}{*}{Software Development} & \textbf{Software agent} & \\
    & Design agent & \\
    & Code agent & \\
    & Test agent & SoftwareDevelopment \\
    & Review agent & SoftwareDevelopment \\
    & Deploy agent & \\
    & Infrastructure agent &  CodeDeployment\\
    & Application agent & CodeDeployment \\
    \bottomrule
    \end{tabular}
\end{table}

\section{Full results of Coordination Mode}
\label{app:full}
\scriptsize
\begin{table}[H]
    \caption{Full End-to-end evaluation of Coordination Mode}
    \label{tab:coord_results_full}
    \centering
    \footnotesize
    \begin{tabular}{p{5cm}p{1.5cm}p{1cm}p{2cm}p{1cm}p{1cm}}
    \toprule
    Setting & Dataset & Overall GSR & Supervisor GSR & User-side GSR & System-side GSR \\
    \midrule 
    \multirow{3}{*}{\shortstack[l]{Supervisor: Sonnet 3.5 (20241022) \\ Specialists: Sonnet 3.5 (20241022)}} & Travel & 0.90 & 0.90 & 1.00 & 0.90 \\
    & Mortgage & 0.90 & 0.93 & 0.93 & 0.90\\
    & Software & 0.90 & 1.00 & 1.00 & 0.90 \\
    \midrule 
    \multirow{3}{*}{\shortstack[l]{Supervisor: Sonnet 3.5 (20241022) \\ Specialists: Sonnet 3.0}} & Travel & 0.87 & 0.90 & 0.90 & 0.87 \\
    & Mortgage & 0.90 & 0.97 & 0.90 & 0.87 \\
    & Software & 0.77 & 0.90 & 0.80 & 0.83 \\
    \midrule
    \multirow{3}{*}{\shortstack[l]{Supervisor: Sonnet 3.5 (20241022) \\ Specialists: Haiku 3.5}} & Travel & 0.80 & 0.87 & 0.83 & 0.83 \\
    & Mortgage & 0.83 & 0.90 & 0.90 & 0.87 \\
    & Software & 0.87 & 0.93  & 0.90 & 0.97\\
    \midrule
    \multirow{3}{*}{Single-agent: Sonnet 3.5 (20241022)} & Travel & 0.60 & -- & 0.67 & 0.67 \\
    & Mortgage & 0.80 & -- & 0.80 & 0.83 \\
    & Software & 0.53 & -- & 0.67 & 0.60 \\
    \bottomrule
    \end{tabular}
\end{table}

\begin{table}[H]
    \caption{Full Latency Performance of Coordination Mode}
    \label{tab:coord_latency_full}
    \centering
    \resizebox{\textwidth}{!}{
    \begin{tabular}{llrrrr}
    \toprule
    Setting & Dataset & \shortstack[r]{Avg. \\ communication \\ overhead \\ per turn (s)} & \shortstack[r]{Avg. \\ user-perceived \\ turn latency \\ per session (s)} & \shortstack[r]{Avg. \\ communications \\ per session} & \shortstack[r]{Avg. output \\ tokens per\\ communication} \\
    \midrule 
    \multirow{3}{*}{\shortstack[l]{Supervisor: Sonnet 3.5 (20241022) \\ Specialists: Sonnet 3.5 (20241022)}} & Travel & 13.75 & 31.46 & 8.63 & 225.88 \\
    & Mortgage & 13.39 & 24.42 & 7.27 & 201.66 \\
    & Software & 35.44 & 168.73 & 7.59 & 373.77 \\
    \midrule 
    \multirow{3}{*}{\shortstack[l]{Supervisor: Sonnet 3.5 (20241022) \\ Specialists: Sonnet 3.0}} & Travel & 15.43 & 42.12 & 7.51 & 276.18 \\
    & Mortgage & 15.97 & 29.90 & 6.69 & 286.21\\
    & Software & 53.48 & 137.31 & 9.11 & 490.78 \\
    \midrule
    \multirow{3}{*}{\shortstack[l]{Supervisor: Sonnet 3.5 (20241022) \\ Specialists: Haiku 3.5}} & Travel & 12.95 & 23.98 & 9.73 & 236.27 \\
    & Mortgage & 11.03 & 18.13 & 6.93 & 202.87 \\
    & Software & 36.65 & 125.31 & 8.07 & 388.79 \\
    \midrule
    \multirow{3}{*}{Single-agent: Sonnet 3.5 (20241022)} & Travel & -- & 14.12 & -- & -- \\
    & Mortgage & -- & 9.12 & -- & -- \\
    & Software & -- & 52.61 & -- & -- \\
    \bottomrule
    \end{tabular}
    }
\end{table}

\section{Human Evaluation Results}
\label{app:human_eval}
\begin{table}[H]
    \caption{The automatic and human evaluation results. Note that these set of experiments were from an intermediate milestone checkpoint of MAC.}
    \label{tab:human_eval_results}
    \centering
    \footnotesize
    \begin{tabular}{llrrrr}
    \toprule
    & Dataset & Overall GSR & Supervisor GSR & User-side GSR & System-side GSR \\
    \midrule 
    \multirow{3}{*}{Automatic Evaluation} & Travel & 0.87 & 0.90 & 0.90 & 0.87 \\
    & Mortgage & 0.90 & 0.97 & 0.97 & 0.90 \\
    & Software & 0.77 & 0.90 & 0.80 & 0.83 \\
    \midrule
    \multirow{3}{*}{Human Evaluation} & Travel & 0.93 & 0.90 & 0.97 & 0.93 \\
    & Mortgage & 0.97 & 0.97 & 0.97 & 0.97 \\
    & Software & 0.73 & 0.87 & 0.80 & 0.73 \\
    \bottomrule
    \end{tabular}
\end{table}

\end{document}